\documentclass{article}

 \usepackage[preprint]{neurips_2026}

\usepackage[utf8]{inputenc} 
\usepackage[T1]{fontenc}    
\usepackage{hyperref}       
\usepackage{url}            
\usepackage{booktabs}       
\usepackage{amsfonts}       
\usepackage{nicefrac}       
\usepackage{microtype}      
\usepackage{xcolor}         
\usepackage{enumitem}
\usepackage{amsmath}
\usepackage{multirow} 
\usepackage{graphicx}
\usepackage{subcaption}
\usepackage{wrapfig}

\newcommand{\modelname}{AME-TS}

\title{\modelname: Anchored Mixture-of-Experts for \\ Time Series Forecasting}

\author{%
  Rui Wang\thanks{Corresponding author: \texttt{rwngamz@amazon.com}} \quad
  Renhao Xue \quad
  Ray Razi \quad
  Huan Song \quad
  Hannah R. Marlowe \\
  Amazon Web Services
}

\begin{document}

\maketitle

\begin{abstract}
Time series forecasting models are increasingly scaled through large Transformer backbones, yet most existing approaches process all series through a shared dense computation path despite substantial heterogeneity in temporal structure. Mixture-of-Experts (MoE) offers a natural alternative by enabling conditional computation, but standard MoE routing leaves expert specialization weakly identified and often unstable during downstream adaptation. We propose \modelname{}, a structure-guided sparse time series foundation model that aligns expert routing with interpretable temporal structure.
\modelname{} first uses a lightweight regime predictor to estimate series-level descriptors, including forecastability, seasonality, trend, and sparsity, and maps them to a soft structural prior over experts. This series-level prior guides token-level routing during training, encouraging structure-aligned specialization. On the GIFT-Eval benchmark, \modelname{} delivers a strong accuracy--efficiency tradeoff across model scales: it substantially outperforms existing time series foundation models at small model scales and remains competitive with the strongest models at larger scales, while activating substantially fewer parameters through sparse routing. We further show that \modelname{} learns more interpretable routing geometry and substantially more stable expert specialization than standard MoE during fine-tuning on the M5 dataset. These results suggest that structure-aware routing is an effective and reliable way to realize the benefits of sparse expert models for time series forecasting.
\end{abstract}

\section{Introduction}
Time series forecasting is a fundamental problem in applications such as retail demand planning \cite{benidis2022deep, wang2025time}, cloud operations\cite{godahewa2021monash, shchur2025fev}, healthcare\cite{morid2023time}, and weather prediction \cite{li2021fourier, wang2020towards}. Time series foundation models have achieved strong zero-shot and transfer performance by scaling Transformer-based architectures over large and diverse collections of time series \cite{ansari2025chronos, cao2025conversational, das2024decoder, ansari2024chronos, shi2025timemoe}. Most of these models, however, process all series through a shared dense computation path. This is an inefficient use of model capacity, because real-world time series differ substantially in the temporal structure relevant for forecasting. A highly seasonal retail series, a strongly trending macroeconomic indicator, and a sparse cloud monitoring signal need not be processed in the same way. These structural differences directly influence which representations and computations are most useful for accurate forecasting.

Mixture-of-Experts (MoE) architectures \cite{cai2025survey} scale neural networks by conditionally activating only a subset of parameters for each input, and have been adopted successfully in both large language models \cite{dai2024deepseekmoe, jiang2024mixtral, fedus2022switch, shen2024mome, shen2024mixtureofexperts} and, more recently, time series forecasting models \cite{liu2024moirai, shi2025timemoe}.  In principle, MoE is a natural fit for forecasting because different time series may benefit from different computational pathways. In practice, however, standard MoE routing leaves expert specialization weakly identified: because experts are permutation-symmetric and therefore do not begin with fixed semantic roles, 
\begin{wrapfigure}{r}{0.50\columnwidth}
    \centering
    \includegraphics[width=\linewidth]{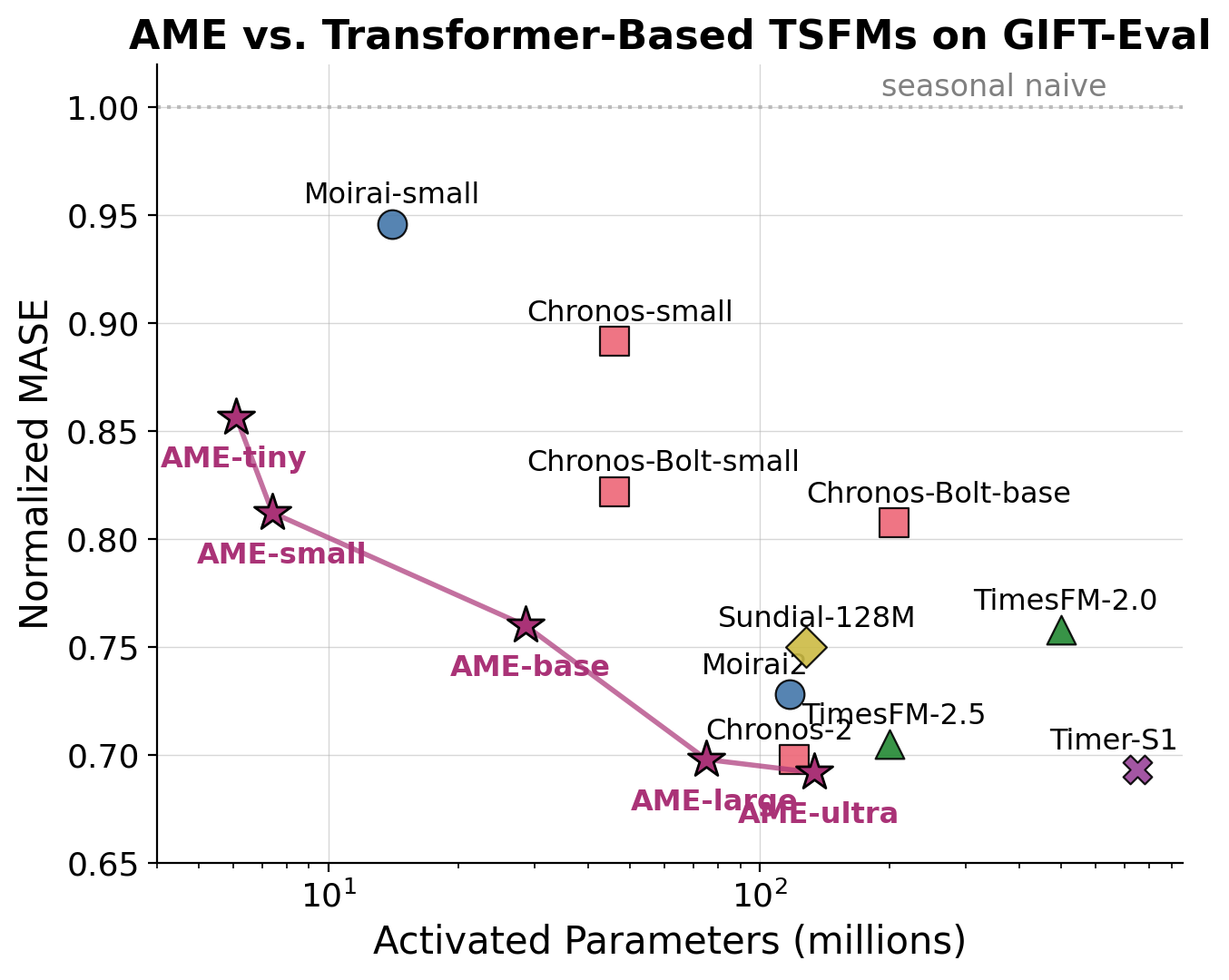}
    \caption{\textbf{MASE vs. activated parameter count on GIFT-Eval}. Each point shows a foundation model or an AME variant, with lower normalized MASE indicating better forecasting performance. \modelname{} achieves a favorable accuracy--efficiency tradeoff across scales, matching or outperforming strong TSFMs while activating substantially fewer parameters through sparse routing.}
    \label{fig:main_results}
    \vspace{-3mm}
\end{wrapfigure}
the router may organize them according to random initialization, optimization dynamics, or fine-tuning dynamics rather than forecasting-relevant temporal structure. 
As a result, expert specialization can be fragile, difficult to interpret, and prone to drift during downstream adaptation. More broadly, recent studies across language, vision, diffusion, and multimodal learning suggest that stronger expert specialization and explicitly guided routing can improve MoE performance and interpretability, further motivating structure-aware routing mechanisms \cite{guo2025advancing, han2026guiding, wei2026routing, min2025guiding, shen2024mixtureofexperts}.

These observations suggest that MoE routing should be guided by meaningful problem
structure, but such guidance must remain flexible in the foundation-model setting.
Hard-coding expert roles or computation paths could limit generalization across
heterogeneous domains. We therefore use temporal structure only as a soft training
signal: it biases expert specialization toward interpretable regimes, while leaving
the learned router free to adapt during broad pretraining, inference, and downstream
transfer.

Time series forecasting is a particularly natural setting for structure-aware routing because it provides interpretable axes along which expert specialization can be organized. Properties such as forecastability \cite{wang2025time, goerg2013forecastable}, seasonality, trend, and sparsity capture distinct aspects of temporal structure that are strongly tied to forecasting performance. 
Motivated by this view, we propose Anchored Mixture-of-Experts (\modelname{}), a structure-guided sparse time series foundation model that uses interpretable temporal descriptors to guide MoE routing.
It first uses a lightweight regime predictor to estimate interpretable temporal descriptors of each input series, including forecastability, seasonality, trend, and sparsity. 
These regime scores are then mapped to a series-level soft prior over experts, which is used during training through a prior-alignment loss to guide token-level routing. 
This encourages structure-aligned specialization while preserving the flexibility of learned routing. 
As a result, \modelname{} breaks the permutation symmetry of standard MoE, leading to more stable, interpretable, and selective expert specialization.

Our contributions are summarized as follows.
\begin{itemize}[leftmargin=20pt]
    \item We propose \modelname{}, a structure-guided sparse time series foundation model that constructs a soft prior over expert usage from interpretable temporal descriptors and introduces a training-only prior-alignment loss to align token-level MoE routing with series-level temporal structure. 
    \item On the GIFT-Eval benchmark \cite{aksu2024gift}, we show that \modelname{} achieves a strong accuracy--efficiency tradeoff, as summarized in Figure~\ref{fig:main_results}: it substantially outperforms existing small-scale time series foundation models and remains competitive with the strongest larger baselines, while using substantially fewer active parameters through sparse routing.
    \item We show that the routing learned by \modelname{} is interpretable, with both routing space and representation space organizing around meaningful temporal regimes.
    \item We demonstrate that anchored routing improves fine-tuning stability: on the M5 dataset \cite{makridakis2022m5}, \modelname{} achieves strong zero-shot and fine-tuned performance while maintaining substantially more stable expert specialization than standard MoE during adaptation.
\end{itemize}

\section{Related Work}

\paragraph{Mixture-of-Experts for Scalable Sequence Modeling}
MoE architectures have become a central approach to scaling neural networks by conditionally activating only a subset of parameters for each input. Early sparse models such as Switch Transformers \cite{fedus2022switch} and later large-scale systems such as Mixtral \cite{jiang2024mixtral} and DeepSeekMoE \cite{dai2024deepseekmoe} show that sparse routing can substantially increase model capacity without proportionally increasing computation. Other variants improve expert utilization through more expressive routing mechanisms or modified expert structures \cite{wu2024multihead, zhang2023samoe}. Although these methods differ in routing design and expert architecture, their primary focus is on scaling efficiency and expert utilization. In most cases, however, expert identities remain emergent, with specialization shaped largely by training dynamics rather than explicit structural guidance. Our work is motivated by the observation that, for time series forecasting, sparse capacity alone is not enough, routing should also align with forecasting-relevant temporal structure.

\paragraph{Guided and Structure-Aware Routing in Mixture-of-Experts}
A growing body of work suggests that MoE performance depends not only on the presence of experts, but also on how expert specialization is induced \cite{guo2025advancing}. Switch-NeRF \cite{mi2023switchnerf} shows that spatially grounded routing improves scene decomposition in neural radiance fields. Routing Matters in MoE \cite{wei2026routing} shows that explicit routing guidance improves expert specialization in diffusion transformers. Guiding the Experts \cite{min2025guiding} uses semantic priors to encourage focused routing in Soft MoE vision models, while \cite{han2026guiding} uses multimodal temporal structure to guide expert allocation. In multimodal large language models, MoME \cite{shen2024mome} and MoVA \cite{zong2024mova} further show that routing aligned with task or modality structure can improve adaptation and generalization. Together, these works suggest that MoE routing benefits from alignment with meaningful problem structure rather than being learned entirely without guidance. More broadly, they support the view that structured routing can improve not only downstream performance, but also the interpretability and stability of expert specialization. Our work follows this line of research, but focuses on time series forecasting. 

\paragraph{Time Series Foundation Models}
Recent progress in time series forecasting has been driven by large pretrained forecasting models. Foundation models such as Chronos \cite{ansari2024chronos, ansari2025chronos}, TimesFM \cite{das2024decoder}, Moirai \cite{woo2024unified}, and the recent xLSTM-based TiRex \cite{auer2025tirex} show that scaling training data and model capacity can yield strong zero-shot and transfer performance across diverse forecasting tasks. More recent work, such as Moirai-MoE \cite{liu2024moirai} and TimeMoE \cite{shi2025timemoe}, has also begun to explore sparse expert architectures for forecasting. 
These works demonstrate the promise of MoE for forecasting, but they do not explicitly use interpretable temporal information to guide expert specialization. In contrast, our approach uses series-level structural descriptors to construct a soft prior over expert assignments. 
\section{Methodology}
\begin{figure*}[t]
\centering
\includegraphics[width=0.9\textwidth]{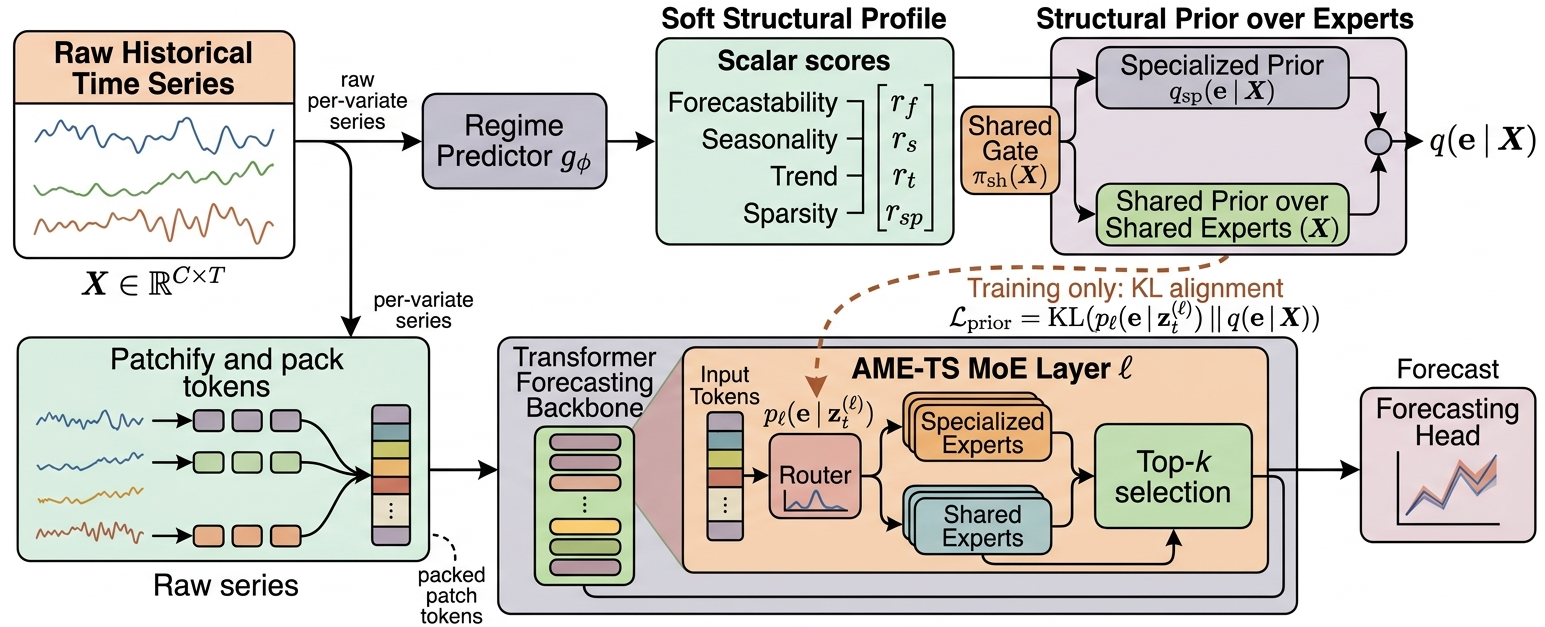}
\caption{Overview of \modelname{}. A regime predictor extracts a soft structural profile from raw time series to construct a structural prior over experts \(q(e \mid X)\), while patchified tokens are processed by a Transformer forecasting backbone with AME-TS MoE layers. Training uses KL alignment between token-level routing and the structural prior; inference uses only the learned router.}
\label{fig:method_overview}
\end{figure*}

\subsection{\modelname{} Overview}
\modelname{} is a structure-guided sparse MoE forecasting model that aligns expert
specialization with interpretable temporal structure. As illustrated in
Figure~\ref{fig:method_overview}, \modelname{} balances a series-level structural
prior with token-level learned routing. The series-level prior summarizes global
temporal properties of the input series, while the token-level router selects experts from local latent representations inside the forecasting backbone. Given an input series, a lightweight regime predictor estimates a soft structural
profile consisting of forecastability, seasonality strength, trend strength, and
sparsity. This profile is mapped to a soft prior over experts, biasing expert
specialization toward interpretable temporal regimes. During training, a
prior-alignment loss encourages token-level routing to align with the series-level
prior while still adapting to fine-grained token-level patterns. During inference, the prior is not explicitly injected and routing is determined only by the learned router.

\subsection{Structural Profile and Expert Prior}
\label{sec:structural_prior}

\paragraph{Structural descriptors.}
A central idea of \modelname{} is that expert routing should reflect the temporal
structure of the time series. We therefore characterize each series using four
complementary structural descriptors: forecastability, seasonality strength, trend
strength, and sparsity. Together, these descriptors capture whether a series is
spectrally regular, periodically structured, directionally changing, or intermittent, providing a compact and interpretable profile for guiding expert specialization.

Forecastability measures how concentrated a series is in the frequency domain, using
the normalized entropy of its power spectral density
\cite{wang2025time, goerg2013forecastable}. High forecastability indicates more
regular spectral structure, while low forecastability corresponds to more noise-like
dynamics. Seasonality strength measures the fraction of non-trend variation explained by the seasonal component in an STL decomposition \cite{cleveland1990stl}. Trend
strength measures the magnitude of normalized linear change across the input window.
Sparsity measures intermittency or repeated-value behavior using the fraction of
non-unique values in the window. These descriptors are not intended to fully
characterize a time series; rather, they provide a small set of interpretable
structural signals that can bias expert routing toward forecasting-relevant regimes.
Detailed mathematical definitions are provided in
Appendix~\ref{app:descriptor_definitions}.

These four descriptors are complementary rather than redundant. Empirically, their
pairwise correlations on the pre-training pool are all moderate, as shown in
Table~\ref{tab:feature_correlation}, indicating that they capture distinct structural axes rather than collapsing to a single structural factor. Our ablation study in Section~\ref{sec:gift_eval} further shows that each descriptor contributes to forecasting performance.

\paragraph{Regime predictor.}
Computing these descriptors analytically for every training sample would be
prohibitively expensive. We therefore compute them on a small subset of series from
the pretraining pool and train a lightweight regime predictor \(g_\phi\) to provide
fast estimates during model training:
\[
g_\phi(X) = [r_{\mathrm{f}}, r_{\mathrm{s}}, r_{\mathrm{t}}, r_{\mathrm{sp}}]
\in [0,1]^4,
\]
where \(X=(x_1,\ldots,x_T)\) denotes a univariate input window of length \(T\)
(for multivariate inputs, descriptors are computed per variate), and \(r_{\mathrm{f}}\), \(r_{\mathrm{s}}\), \(r_{\mathrm{t}}\), and
\(r_{\mathrm{sp}}\) denote the predicted scores for forecastability, seasonality
strength, trend strength, and sparsity, respectively. The regime predictor is trained
separately and kept frozen during \modelname{} training. Architectural and training
details are provided in Appendix~\ref{app:regime_predictor}.

\paragraph{Specialized and shared experts.}
Using the regime predictor, we construct a prior over experts that biases
routing toward forecasting-relevant temporal structure. We partition the expert set
into two groups: \emph{specialized experts} \(\mathcal{E}_{\mathrm{sp}}\), which are
associated with the four structural descriptors, and \emph{shared experts}
\(\mathcal{E}_{\mathrm{sh}}\), which provide fallback capacity when
structural profile is weak, mixed, or ambiguous.

We define a fixed anchor distribution \(q_{\mathrm{anchor}}(e \mid d)\) over
specialized experts by assigning each descriptor \(d\) to a subset of specialized
experts. In practice, \(q_{\mathrm{anchor}}(e \mid d)\) is uniform over the experts
assigned to descriptor \(d\) and zero elsewhere. When the number of specialized
experts exceeds the number of descriptors, experts are distributed as evenly as
possible across descriptors. The regime-induced prior over specialized experts is
defined as
\[
q_{\mathrm{sp}}(e \mid X) \propto
\sum_{d=1}^{D} q_{\mathrm{anchor}}(e \mid d)\, g_{\phi}(d \mid X),
\qquad e \in \mathcal{E}_{\mathrm{sp}}.
\]
Intuitively, higher regime
scores place more prior mass on the experts anchored to those descriptors. Because
the structural profile is soft, the resulting prior can simultaneously favor multiple
expert groups when a series exhibits mixed structure.

Not all series admit strong descriptor-specific specialization. Some exhibit weak
structural signals, while others lie between multiple regimes. To handle such cases,
we allocate part of the prior mass to a shared expert pool through a shared gate
\(\pi_{\mathrm{sh}}(X)\in[0,1]\), which increases when the regime profile is weak or
uncertain. Let
\[
H(X)=\frac{1}{D}\sum_{d=1}^{D} h\!\left(g_\phi(d\mid X)\right),
\qquad
S(X)=\max_{d \in \{1,\ldots,D\}} g_\phi(d\mid X),
\]
where \(h\) is the binary entropy. We define the shared gate as
\[
\pi_{\mathrm{sh}}(X)=\big(1-S(X)\big)\,
\sigma\!\left(\alpha H(X)-b\right),
\]
where \(\sigma(\cdot)\) is the sigmoid function, and \(\alpha\) and \(b\) are
learnable or fixed parameters. This form reflects two intuitions: shared experts
should receive more mass when the structural profile is uncertain, and when no
descriptor is strongly activated. The factor \(1-S(X)\) suppresses shared mass when
at least one descriptor is strongly activated, while the entropy term increases shared mass when the predicted descriptor scores are diffuse. 

The final prior over all experts is given by
\[
q(e \mid X) =
\begin{cases}
\displaystyle \frac{\pi_{\mathrm{sh}}(X)}{|\mathcal{E}_{\mathrm{sh}}|},
& e \in \mathcal{E}_{\mathrm{sh}}, \\[8pt]
\displaystyle \big(1-\pi_{\mathrm{sh}}(X)\big)\, q_{\mathrm{sp}}(e \mid X),
& e \in \mathcal{E}_{\mathrm{sp}}.
\end{cases}
\]
This prior is soft and interpretable: it breaks the permutation symmetry of standard
MoE routing by anchoring expert preferences to meaningful regimes, while allowing
uncertain cases to be absorbed by shared experts. Rather than hard-assigning experts,
it provides a series-level structural bias that is later used to guide token-level
routing.

\subsection{Training with Prior Alignment}
\label{sec:prior_alignment}

The key training mechanism of \modelname{} is a prior-alignment loss that transfers
series-level temporal structure into token-level sparse routing. The structural prior
\(q(e \mid X)\) defined in Section~\ref{sec:structural_prior} is constructed at the
series level, whereas routing in the MoE layers operates at the token level. Given a
token representation \(z_t^{(\ell)}\) at layer \(\ell\),  where \(t\) indexes patch tokens, the router produces logits
\[
s_\ell(z_t^{(\ell)}) = W_r^{(\ell)} z_t^{(\ell)},
\]
which define a token-level routing distribution
\[
p_\ell(e \mid z_t^{(\ell)}) =
\mathrm{softmax}\big(s_\ell(z_t^{(\ell)})\big).
\]
As in standard sparse MoE, only the top-\(k\) experts are activated per token. Rather
than injecting the structural prior directly into the router logits, we use it only
during training as a soft signal that encourages alignment between token-level routing
and series-level structure. 
Specifically, we introduce a Kullback--Leibler (KL) regularization term:
\[
\mathcal{L}_{\mathrm{prior}}=
\frac{1}{N_L}\sum_{\ell=0}^{N_L-1}\lambda_\ell\;
\mathbb{E}_{t}\left[
\mathrm{KL}\big(p_\ell(e\mid z_t^{(\ell)}) \,\|\, q(e\mid X)\big)
\right],
\]
where \(N_L\) is the number of Transformer layers. We use the forward KL divergence
\(\mathrm{KL}(p \,\|\, q)\), which encourages the token-level router to remain
consistent with the soft structural prior. This prior-alignment loss is the main
mechanism that transfers interpretable series-level structure into token-level sparse
routing. It encourages experts to develop stable structural roles, while still
allowing the learned router to adapt to fine-grained token-level patterns.

\paragraph{Layer-wise prior weighting.}
To control where specialization emerges in the network, we vary the strength of prior
regularization across layers. We adopt a linearly increasing schedule:
\[
\lambda_\ell = \lambda_{\max}\frac{\ell}{N_L-1},
\]
where \(\ell \in \{0,\dots,N_L-1\}\) is the layer index. This design applies minimal
regime pressure to early layers, allowing them to learn shared representations, while encouraging deeper layers to specialize according to the regime prior. This reflects the intuition that high-level structural information is more useful in later stages of computation.

\paragraph{Orthogonality loss.}
When multiple experts are associated with the same descriptor, we further include an
orthogonality loss to promote diversity among their outputs:
\[
\mathcal{L}_{\mathrm{ortho}} =
\mathbb{E}_{i \neq j} \left[ \left| \langle h_i, h_j \rangle \right| \right],
\]
where \(h_i\) and \(h_j\) are outputs of co-activated experts within the same group.
In practice, we find that the prior-alignment term substantially mitigates expert
collapse and stabilizes routing, while the orthogonality loss provides additional but
smaller gains, especially when using larger expert pools.

\subsection{Forecasting Backbone and Prediction Loss}
\label{sec:backbone_loss}

\modelname{} is built on an encoder-only Transformer forecasting backbone. Given an
input window, each variate is treated as a univariate series, partitioned into
non-overlapping patches, and projected into latent token representations. For
multivariate inputs, tokens from all variates are packed into a single sequence with
variate identity embeddings and processed by a stack of Transformer encoder layers.
Forecasting is formulated as masked prediction over the forecast horizon: historical
tokens are provided as observed input, while tokens in the future horizon are masked.
The encoder outputs at these masked positions are then used to predict the target
values. We replace the dense feed-forward layers in each encoder block with MoE
layers, while keeping the rest of the backbone unchanged. This backbone design follows
the packed-sequence formulation of \cite{woo2024unified}.

The primary forecasting objective \(\mathcal{L}_{\mathrm{task}}\) is a masked
prediction loss over the forecast horizon. Given predicted values \(\hat{y}\) and
ground-truth targets \(y\), we use an \(\ell_1\) loss:
\[
\mathcal{L}_{\mathrm{task}} =
\mathbb{E}\left[\|\hat{y} - y\|_1\right].
\]
The training objective combines the prediction loss with the prior-alignment
and orthogonality losses:
\[
\mathcal{L} =
\mathcal{L}_{\mathrm{task}}
+ \lambda_{\mathrm{prior}}\mathcal{L}_{\mathrm{prior}}
+ \lambda_{\mathrm{ortho}}\mathcal{L}_{\mathrm{ortho}}.
\]

\section{Experiment}

\begin{table*}[t]
\centering
\begin{minipage}[t]{0.55\textwidth}
\centering
\small
\setlength{\tabcolsep}{4pt}
\renewcommand{\arraystretch}{0.97}
\captionof{table}{\textbf{Summary GIFT-Eval results.} Scores are geometric means over 97 tasks normalized by the Seasonal Naive baseline; lower is better. Full results across all four metrics are provided in Appendix Table~\ref{app_tab:main_results}. $^\dagger$ Active parameters per token due to sparse MoE routing.}
\label{tab:main_results}
\begin{tabular}{lccc}
\toprule
\textbf{Model} & \textbf{\#Params} & \textbf{MASE $\downarrow$} & \textbf{RMSE $\downarrow$} \\
\midrule
Moirai-Small \cite{woo2024unified}         & 14M               & 0.946 & 1.056 \\
Chronos-Small \cite{ansari2024chronos}     & 46M               & 0.892 & 0.866 \\
Chronos-2 \cite{ansari2025chronos}         & 120M              & 0.698 & 0.704 \\
Sundial-128M \cite{liu2025sundial}         & 128M              & 0.750 & 0.739 \\
TimesFM-2.5 \cite{das2024decoder}          & 231M              & 0.705 & 0.694 \\
Moirai2 \cite{liu2025moirai}               & 300M              & 0.728 & 0.739 \\
\midrule
\modelname{} Tiny                          & 14M$^\dagger$6M  & 0.856 & 0.808 \\
\modelname{} Small                         & 30M$^\dagger$7M  & 0.812 & 0.767 \\
\modelname{} Base                          & 62M$^\dagger$29M & 0.765 & 0.737 \\
\modelname{} Large                         & 162M$^\dagger$75M & 0.700 & 0.690 \\
\modelname{} Ultra                         & 540M$^\dagger$133M & \textbf{0.692} & \textbf{0.687} \\
\bottomrule
\end{tabular}
\end{minipage}
\hfill
\begin{minipage}[t]{0.43\textwidth}
\centering
\vspace{0.9mm}
\includegraphics[width=\linewidth]{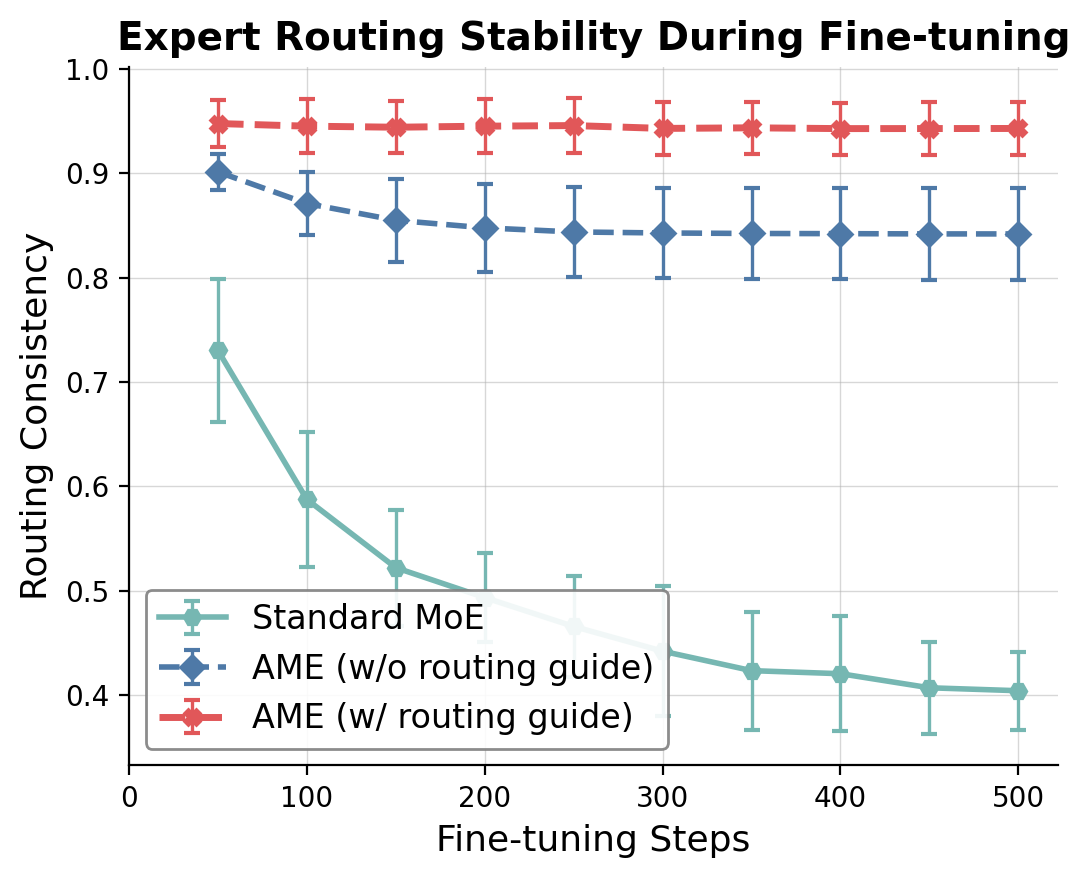}
\captionof{figure}{\textbf{Routing stability during fine-tuning on M5.} \modelname{} maintains substantially more stable expert specialization than standard MoE, and routing guidance further improves stability during adaptation.}
\label{fig:routing_stability}
\end{minipage}
\end{table*}

\subsection{Experimental Setup}
We pre-train \modelname{} on a pre-training pool spanning 96 dataset configurations across 8 domains (Table~\ref{tab:pretrain_data}), comprising approximately 3.5 million individual time series and 18 billion observations, with frequencies ranging from seconds to yearly. A detailed breakdown is provided in Appendix~\ref{app:data_pool}. In addition to the foundation-model setting, we also evaluate \modelname{} in a dataset-specific setting, where a smaller model is trained independently for each GIFT-Eval task.

We train using AdamW with a linear decay learning-rate schedule and linear warmup. Training is performed on a single p4 instance with 8 A100 GPUs. For each task, we sweep over eight context lengths, select the best setting on the validation set, and use that context length for the final test-set evaluation. Following the official protocol \cite{aksu2024gift}, we report four metrics, MASE, sMAPE, MAE, and RMSE, computed per series and then aggregated across the 97 tasks after normalization by the Seasonal Naive baseline. Detailed architecture configurations for all \modelname{} variants are provided in Appendix Table~\ref{tab:ame_family}, and additional training details are provided in Appendix~\ref{app:hyperparameter}. We will release code and implementation details upon publication.

\subsection{GIFT-Eval Results and Ablation} \label{sec:gift_eval}

\paragraph{Main results}
We evaluate \modelname{} on the GIFT-Eval benchmark, which comprises 97 forecasting tasks spanning diverse datasets, frequencies, and prediction horizons. We compare against published time series foundation models at three parameter scales, as well as task-specific per-dataset models. For each scale group, we report the top published models at that scale from the official leaderboard. Although Moirai-MoE \cite{liu2024moirai} and TimeMoE \cite{shi2025timemoe} are relevant forecasting MoE models in the literature, they do not appear as directly comparable named entries on the public GIFT-Eval leaderboard at the time of writing, so our main comparison follows the top publicly listed leaderboard baselines.

Figure~\ref{fig:main_results} and Table~\ref{tab:main_results} summarize the main
GIFT-Eval comparison. Figure~\ref{fig:main_results} highlights the
accuracy--efficiency tradeoff in terms of MASE versus activated parameter count,
while Table~\ref{tab:main_results} reports the corresponding MASE and RMSE values.
\modelname{} achieves the strongest MASE and RMSE among the listed models, with
\modelname{} Ultra reaching 0.692 MASE and 0.687 RMSE while activating 133M
parameters per token. The gains are especially pronounced at smaller scales:
\modelname{} Small outperforms both Moirai-Small and Chronos-Small while activating
only 7M parameters per token. Full results across all four metrics, including sMAPE and MAE, as well as task-specific per-dataset model comparisons, are provided in Appendix
Table~\ref{app_tab:main_results}.

\paragraph{Ablation study}
Beyond this main comparison, we conduct controlled ablations to isolate the contribution of routing design, as shown in Table~\ref{tab:ablation}. Replacing a dense model with standard MoE yields modest improvements, confirming that sparse expert capacity alone is beneficial. Introducing regime-aware routing yields further gains, showing that the improvement comes not only from increased capacity, but from better alignment between routing decisions and temporal structure. Among prior-integration strategies, KL-guided routing consistently outperforms additive prior injection, indicating that soft alignment is more effective than direct prior injection. Forward and reverse KL achieve similar performance, with forward KL showing slightly more consistent gains across tasks.

We further study the contribution of individual regime features by dropping one feature at a time. Removing any single feature degrades performance, confirming that the four regime descriptors provide complementary routing signals. The largest drops are observed when forecastability or sparsity is removed, suggesting that these features provide especially informative cues for expert allocation. Taken together, these ablations show that the gains of \modelname{} arise from structure-aligned routing rather than sparse capacity alone.


\begin{figure}[t]
\centering
\begin{subfigure}[t]{0.48\columnwidth}
    \centering
    \includegraphics[width=\linewidth]{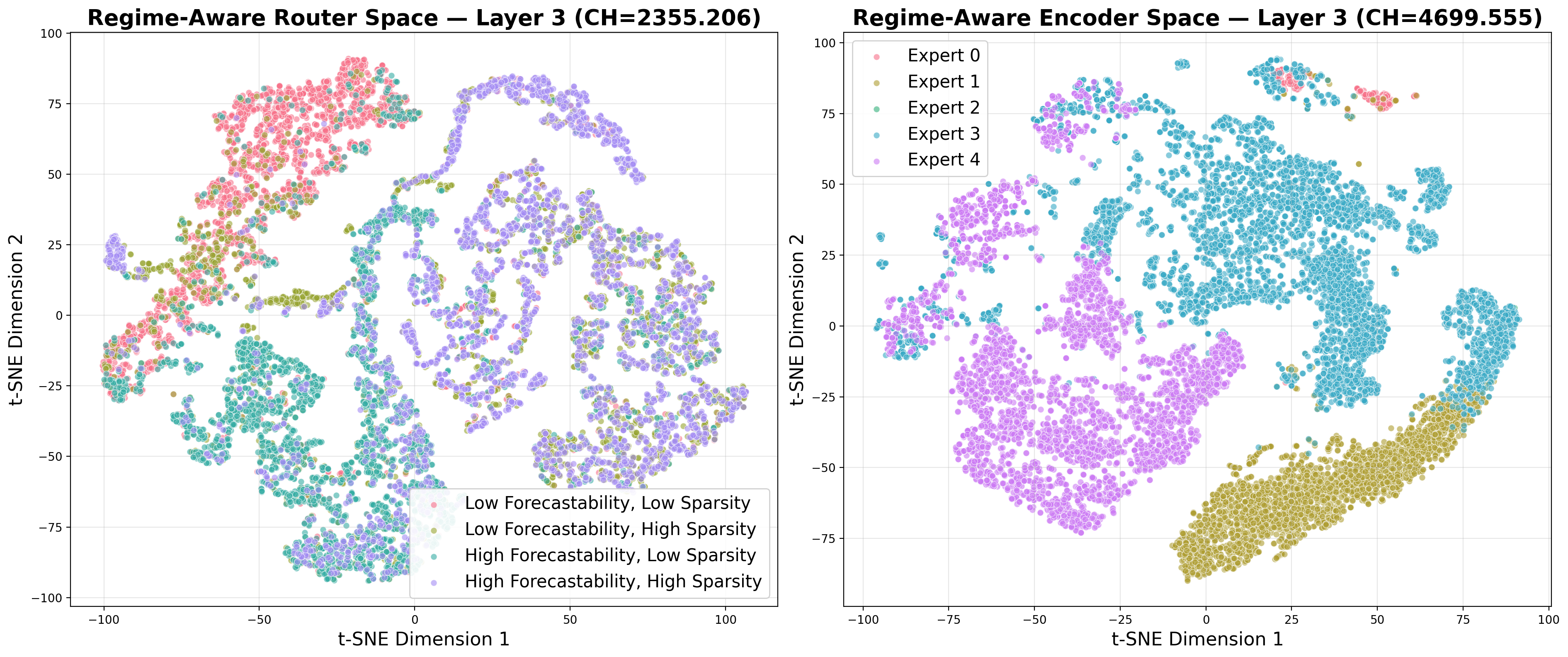}
    \caption{Regime-Aware (AME)}
\end{subfigure}
\hspace{0.01\columnwidth}
\begin{subfigure}[t]{0.48\columnwidth}
    \centering
    \includegraphics[width=\linewidth]{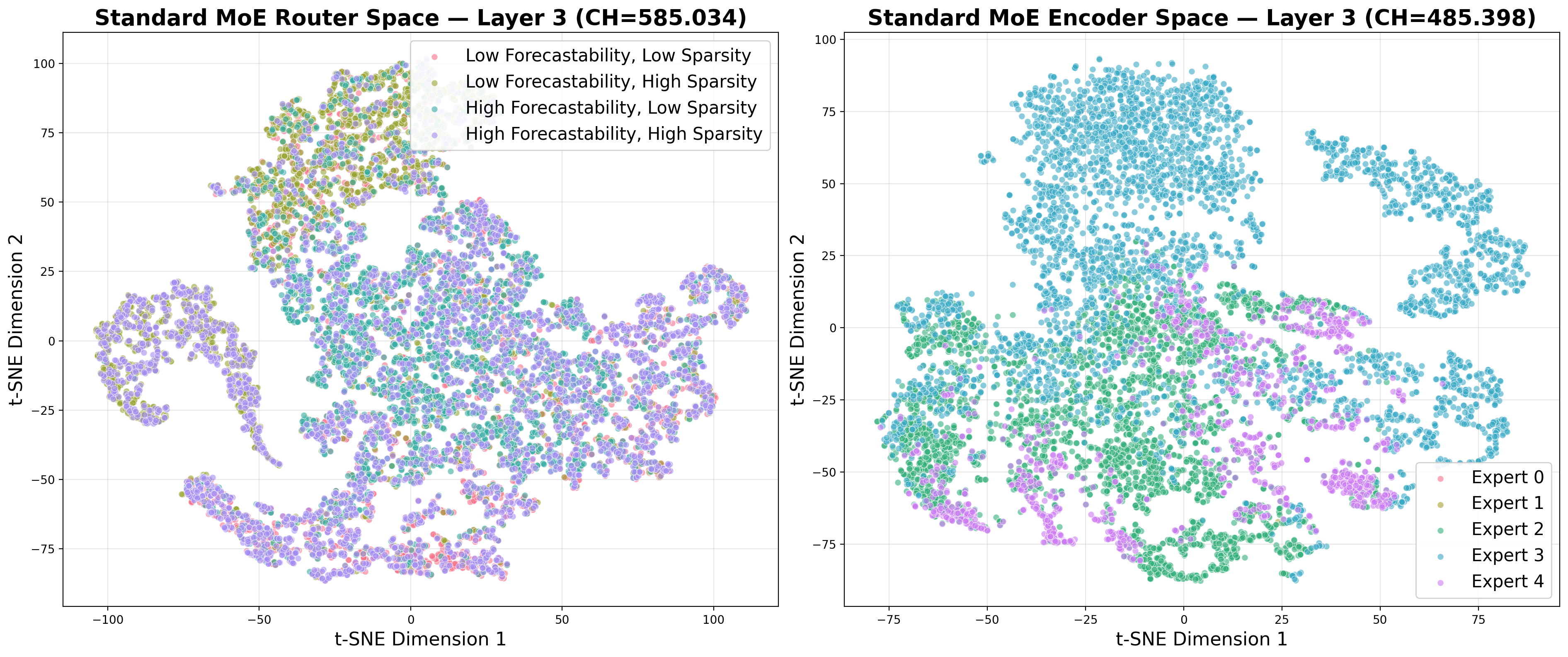}
    \caption{Standard MoE}
\end{subfigure}
\caption{t-SNE visualizations comparing \modelname{} and standard MoE at the same layer. Each subfigure contains two panels: router space on the left and encoder space on the right. In the router space, points are colored by regime profiles derived from forecastability and sparsity labels. In the encoder space, points are colored by top-1 expert assignment. \modelname{} yields clearer regime-aligned structure in router space and substantially stronger expert separation in encoder space, as also reflected by the higher Calinski--Harabasz (CH) scores reported in the subplot titles.}
\label{fig:router_tsne}
\end{figure}
\subsection{Routing Interpretability and Representation Analysis}\label{sec:routing_stability}

We evaluate whether \modelname{} learns routing patterns and internal representations that align with interpretable temporal structure. To this end, we compare \modelname{} against a standard MoE model with the same backbone and expert architecture but without regime-aware routing. From the same layer in both models, we extract router logits as routing embeddings, encoder representations, top-1 expert assignments, and regime predictions from the regime predictor.

To visualize the learned spaces, we project both routing embeddings and encoder representations into two dimensions using t-SNE. In the router space, points are colored by regime profiles obtained by thresholding the regime predictor outputs. In Figure~\ref{fig:router_tsne}, we focus on binary labels derived from forecastability and sparsity, which yields four regime groups. In encoder space, points are colored by top-1 expert assignment. This lets us separately examine whether routing geometry aligns with temporal structure and whether encoder representations organize according to expert specialization.

Figure~\ref{fig:router_tsne} shows a clear qualitative difference between \modelname{} and standard MoE. In the router space, \modelname{} forms more coherent regions associated with different regime labels, whereas standard MoE exhibits substantially greater mixing across groups. In the encoder space, the contrast is even stronger: \modelname{} produces sharply separated expert-specific regions, while standard MoE yields weaker and more entangled clusters. These patterns suggest that the regime prior not only affects routing decisions directly, but also reshapes the representation geometry learned by the backbone.

We quantify this effect using the Calinski--Harabasz (CH) index \cite{wang2019improved}, which measures cluster compactness and separation. Higher CH values indicate better-defined clusters. As reported in the subplot titles of Figure~\ref{fig:router_tsne}, \modelname{} achieves substantially higher CH scores than standard MoE in both router and encoder space. The gain is especially pronounced in the encoder space, where expert assignments under \modelname{} form highly separable regions. This indicates that regime-aware routing leads to more structured routing behavior and stronger expert specialization.

Overall, these results suggest that the gains of \modelname{} are not merely due to increased sparse capacity. Instead, the regime prior induces a more interpretable routing geometry, in which series with similar temporal structure are routed more consistently, and expert-specific computation becomes more clearly separated in representation space.

\subsection{Zero-Shot and Fine-Tuning on M5}
We evaluate \modelname-Base on the M5 Walmart dataset \cite{makridakis2022m5}, which is not included in pre-training. M5 contains 30,490 daily retail time series organized into 12 hierarchical levels and is evaluated using WRMSSE over a 28-day forecasting horizon. Because many recent forecasting foundation models include M5 in pre-training, we avoid direct zero-shot comparison to such models and instead report results against the first-place M5 competition result as a task-specific reference.

\begin{table*}[t]
\centering
\scriptsize
\tabcolsep=3.5pt
\caption{WRMSSE on the M5 dataset across all 12 hierarchical levels. Lower is better. ``Rank 1'' denotes the first-place M5 competition result, reported here as a task-specific reference.}
\label{tab:m5_levels}
\begin{tabular}{cccccccccccccc}
\toprule
Model & Tot & St & Str & Cat & Dept & St-Cat & St-Dept & Str-Cat & Str-Dept & Prod & Prod-St & Prod-Str & Average \\
\midrule
M5 Rank 1  \cite{makridakis2022m5}           & 0.199 & 0.310 & 0.400 & 0.277 & 0.365 & 0.390 & 0.474 & 0.480 & 0.573 & 0.966 & 0.929 & 0.884 & 0.520 \\
\modelname{}  (Zero-shot)       & 0.367 & 0.470 & 0.536 & 0.424 & 0.553 & 0.517 & 0.620 & 0.606 & 0.684 & 0.851 & 0.860 & 0.867 & 0.613 \\
\modelname{}  (Fine-tuned)      & 0.226 & 0.357 & 0.430 & 0.318 & 0.441 & 0.419 & 0.521 & 0.502 & 0.579 & 0.754 & 0.781 & 0.765 & \textbf{0.506} \\
\bottomrule
\end{tabular}
\end{table*}

We first evaluate zero-shot forecasting without any task-specific adaptation. Even without fine-tuning, \modelname{} outperforms the competition winner on all three item-level aggregations (Prod, Prod-St, Prod-Str). This is especially notable because lower-level M5 series are more heterogeneous and often exhibit higher sparsity and less regular temporal structure. These results suggest that regime-aware routing transfers effectively to an unseen retail dataset with diverse structural characteristics. 

We then fine-tune \modelname{} on M5 training data for 20K steps with a batch size of 16, incorporating a day-of-week calendar covariate as a second input variate. Because the model treats each variate as a separate token sequence sharing the same time index, the day-of-week signal is fully observed at both context and forecast positions. Fine-tuned \modelname{} achieves an average WRMSSE of 0.506, improving over the first-place M5 result (0.520), which relied on ensembles of gradient-boosted models with extensive hand-crafted features and hierarchical reconciliation. The gains are strongest at the item level, where \modelname{} reduces WRMSSE by 13--22\% relative to Rank~1. Fine-tuned \modelname{} still falls short of Rank~1 on some higher aggregation levels, where strong seasonality and hierarchical reconciliation favor heavily engineered task-specific pipelines. Nevertheless, the item-level improvements are large enough to yield a lower average WRMSSE overall.

\subsection{Routing Stability During Fine-Tuning}
We evaluate whether expert specialization remains stable during fine-tuning on the M5 dataset. We compare \modelname{} against two baselines with identical architectures: standard MoE (no prior) and an ablated variant of \modelname{} in which routing guidance is removed during fine-tuning.

We define routing consistency as the agreement between current routing decisions and those at the initial checkpoint. For a fixed probe set $\mathcal{D}_{\text{probe}}$, let $e_{i,t,\ell}^{(0)}$ and $e_{i,t,\ell}^{(k)}$ denote the top-1 expert assigned to token $t$ of series $i$ at layer $\ell$ at the initial and $k$-th fine-tuning step, respectively. We compute
\[
\mathrm{RC}(k)
=
\frac{1}{|\mathcal{S}|}
\sum_{(i,t,\ell)\in \mathcal{S}}
\mathbf{1}\!\left[
e_{i,t,\ell}^{(k)} = e_{i,t,\ell}^{(0)}
\right],
\]
where $\mathcal{S}$ denotes the set of tracked series-token-layer tuples derived from $\mathcal{D}_{\text{probe}}$. Higher values indicate more stable expert specialization. We construct a fixed probe set of 1,000 time series sampled across all hierarchy levels of M5 to capture diverse temporal behaviors.

Figure~\ref{fig:routing_stability} shows routing consistency over fine-tuning steps. Confidence intervals are computed over five independent fine-tuning runs with learning rates sampled from $[1\times10^{-5}, 5\times10^{-5}]$. \modelname{} with regime-aware routing guidance maintains consistently high stability throughout training, indicating that expert roles are preserved during fine-tuning under the guidance of the regime predictor.

Notably, the ablated variant without routing guidance also remains stable, with only a modest decrease from $\sim$0.90 to $\sim$0.84. This suggests that \modelname{} learns meaningful and well-structured expert specialization during pre-training, which is largely preserved even without explicit guidance during fine-tuning. In contrast, standard MoE exhibits substantial drift, with consistency dropping from $\sim$0.73 to $\sim$0.40, indicating that expert assignments are continuously reconfigured during fine-tuning.

These results show that \modelname{} substantially mitigates routing instability in MoE during fine-tuning. By learning regime-aligned routing, \modelname{} produces expert specialization that remains coherent during adaptation, avoiding the collapse and drift observed in standard MoE.

\section{Discussion}
\modelname{} shows that aligning expert routing with temporal structure can improve both forecasting efficiency and adaptation stability in Mixture-of-Experts for time series forecasting. On GIFT-Eval, it delivers a strong accuracy--efficiency tradeoff across model scales, with especially large gains at small model sizes and competitive performance at larger scales while activating substantially fewer parameters. On M5, it shows strong transfer in both zero-shot and fine-tuned settings, with the largest gains appearing at lower hierarchical levels where the series are more diverse, sparse, and weakly structured. Beyond predictive accuracy, \modelname{} learns structured expert specialization that remains substantially more stable than standard MoE during fine-tuning. Together, these results suggest that domain-informed routing priors can make sparse forecasting models both more reliable and more interpretable. Importantly, this does not require hard-coding expert assignments or forecasts. Temporal descriptors provide a structural bias for organizing sparse capacity during training, while the learned router retains the flexibility needed for broad pretraining and downstream adaptation. A limitation of the current work is that \modelname{} focuses on historical time series
inputs, while future dynamics may also depend on external context such as text,
events, or metadata. A natural next direction is multimodal forecasting, where
structure-aware routing could combine temporal descriptors with such external
context to further improve forecasting performance and interpretability.
\newpage
\bibliographystyle{plain}
\bibliography{neurips_2026.bib}

\newpage
\appendix
\section{Additional Experimental and Implementation Details}

\subsection{Model Architecture Details} \label{app:model_details}
We evaluate five \modelname{} variants at different scales: Tiny, Small, Base, Large, and Ultra. Their detailed configurations are reported in Table~\ref{tab:ame_family}. In the dataset-specific setting, model size is selected separately for each dataset and is generally much smaller than in the foundation-model setting.
\subsection{Structural Descriptor Definitions}
\label{app:descriptor_definitions}

We compute four structural descriptors for each input series: forecastability,
seasonality strength, trend strength, and sparsity. These descriptors provide
complementary signals for constructing the structural prior used in \modelname{}.

\paragraph{Forecastability.}
Forecastability measures how predictable a series is from its frequency-domain
structure. Following prior work \cite{wang2025time, goerg2013forecastable}, we
quantify it using the entropy of the normalized power spectral density. Given an
input series $X=(x_0,x_1,\ldots,x_{T-1})$, let $\widetilde{X}=\mathrm{Detrend}(X)$
denote its detrended version, and let $p_i$ denote the normalized power of the
$i$-th frequency bin of $\widetilde{X}$. The spectral entropy is
\[
H_a(\widetilde{X}) = -\sum_i p_i \log_a p_i,
\]
and the forecastability score is
\[
\mathrm{Forecastability}(X)
=
1 - \frac{H_a(\widetilde{X})}{\log_a N_f},
\]
where $N_f$ is the number of frequency bins.

\paragraph{Seasonality strength.}
We quantify seasonality strength using STL decomposition \cite{cleveland1990stl}.
Each series is decomposed into trend, seasonal, and remainder components, where the
seasonal period is determined from the dominant Fast Fourier Transform peak frequency.
Let $S$ and $R$ denote the seasonal and residual components, respectively. We define
\[
\mathrm{SeasonalityStrength}(X)
=
1 - \frac{\mathrm{Var}(R)}{\mathrm{Var}(S+R)}.
\]

\paragraph{Trend strength.}
Trend strength measures the magnitude of linear change across the input window. We
first min--max normalize the series to $[0,1]$, fit a linear regression, and let
$\hat{\beta}$ denote the fitted slope. We define
\[
\mathrm{TrendStrength}(X)
=
\min(1, |\hat{\beta}|T).
\]
This measures the directional change over the input window relative to the series
range.

\paragraph{Sparsity.}
Sparsity captures how intermittent or repeated-value dominated a series is. We define
\[
\mathrm{Sparsity}(X)
=
1 - \frac{N_{\mathrm{unique}}(X)}{T},
\]
where $N_{\mathrm{unique}}(X)$ is the number of unique values in the input window.

\subsection{Regime Predictor Architecture and Training}
\label{app:regime_predictor}

The regime predictor $g_\phi$ is trained separately from the forecasting model and kept frozen during \modelname{} training. Its role is to map a raw input series to a soft structural profile over the four structural descriptors: forecastability, seasonality strength, trend strength, and sparsity. Because these properties are not mutually exclusive, $g_\phi$ predicts them independently rather than assigning each series to a single discrete class. The resulting output is a soft structural profile that summarizes the forecasting-relevant characteristics of the input series.

The predictor is trained on a subset of the same pre-training pool used for the forecasting model. Computing the analytical structural descriptors for every series in the full pre-training pool would be prohibitively expensive, so we instead construct a sampled training set by drawing approximately 98,910 random crops. For each crop, ground-truth structural descriptor targets are computed using the analytical feature pipeline described in Section~\ref{sec:structural_prior}.

Rather than using a single multi-output network, we train four separate single-feature predictors, one for each structural descriptor: forecastability, seasonality strength, trend strength, and sparsity. Each predictor has approximately 450K parameters, for a total of roughly 1.8M parameters across all four predictors.

The architecture consists of a multi-scale 1D CNN encoder with three parallel branches using kernel sizes 5, 11, and 21. Each branch contains two Conv1d layers with GroupNorm, GELU activations, and MaxPool(2). In parallel, we apply a self-attention block to the first-scale feature map, consisting of LayerNorm, 4-head MultiheadAttention, and mean pooling. The pooled outputs from the three CNN branches and the attention block are concatenated into a 512-dimensional representation, which is fed to an MLP head of the form $512 \rightarrow 128 \rightarrow 64 \rightarrow 1$, with GELU activations, dropout of 0.1 after the first hidden layer, and a sigmoid output in $[0,1]$.

The four structural descriptors exhibit heterogeneous label distributions. For example, forecastability is often concentrated in a relatively narrow range, whereas sparsity spans a much broader portion of $[0,1]$. Training directly on raw targets can therefore bias the predictors toward densely populated regions of the label distribution. To mitigate this, we apply per-feature quantile normalization to the regime targets before training. For each feature, target values are mapped to their empirical ranks within the sampled training pool, yielding approximately uniform targets on $[0,1]$. This encourages the predictor to use the full target range rather than focusing disproportionately on highly concentrated regions. The same training quantiles are used to normalize validation targets.

We train the regime predictors using mean squared error loss on the quantile-normalized targets. Optimization uses AdamW with learning rate $10^{-3}$, weight decay $10^{-4}$, cosine annealing for 100 epochs, batch size 256, and early stopping with patience 15 based on validation MSE.

\subsection{Pre-training Data Composition} \label{app:data_pool}
We pre-train \modelname{} on a heterogeneous corpus assembled from three public sources. First, we draw a subset of datasets from the LOTSA archive~\cite{woo2024unified}, which contains more than 170 datasets totaling 27 billion observations; our subset represents less than 20\% of the full archive and spans traffic, weather, energy, retail, healthcare, web/cloud, and economics. Second, we incorporate several unique datasets from the Chronos pre-training corpus~\cite{ansari2024chronos}, including a one-million-series synthetic KernelSynth corpus and several real-world datasets not otherwise available in LOTSA. Third, we generate a synthetic dataset of 87,000 time series from diverse parametric models, including seasonal, trend, sparse, and noise processes, with known structural descriptor labels across four dimensions (forecastability, seasonality, trend, sparsity) to encourage expert specialization during routing. We explicitly exclude the M5 retail hierarchy from pre-training, reserving it for downstream fine-tuning and evaluation, and we verify that our corpus contains no overlap with the held-out test windows of standard forecasting benchmarks, so all reported results are free of test-data leakage. In total, the pre-training pool spans 96 dataset configurations across 8 domains (Table~\ref{tab:pretrain_data}), comprising approximately 3.5 million individual time series and 18 billion observations, with frequencies ranging from 10-second to yearly. We use domain-balanced sampling with per-dataset weights to prevent large datasets (e.g., buildings\_900k, kernel\_synth\_1m) from dominating training.

\begin{table}[t]
\centering
\small
\caption{Pre-training data composition (96 configurations, 8 domains).}
\label{tab:pretrain_data}
\begin{tabular}{@{}lccc@{}}
\toprule
Domain & \#Datasets & \#Series & \#Obs. \\
\midrule
Econ/Fin       & 15 & 104K  & 28M \\
Energy         & 23 & 1.80M & 16.16B \\
Transport      & 16 & 5K    & 148M \\
Weather/Climate& 17 & 70K   & 65M \\
Sales/Retail   &  7 & 246K  & 227M \\
Healthcare     &  6 & 1K    & 0.2M \\
Web/Cloud      & 10 & 207K  & 371M \\
Synthetic      &  2 & 1.09M & 1.04B \\
\midrule
\textbf{Total} & \textbf{96} & \textbf{3.52M} & \textbf{18.04B} \\
\bottomrule
\end{tabular}
\end{table}

\subsection{Additional Training Hyperparameters}\label{app:hyperparameter}
For pre-training, we optimize \modelname{} using AdamW with $\beta_1=0.9$, $\beta_2=0.98$, and weight decay $0.01$. We use a peak learning rate of $5\times10^{-4}$ with a cosine-with-restarts schedule over three cycles and a warmup of 5,000 steps. Training runs for 200 epochs with 2,000 steps per epoch, for a total of 400K optimization steps. We use a batch size of 32 per GPU on 8 GPUs, giving an effective batch size of 256, and apply gradient clipping with maximum norm 2.0. During masked forecasting pre-training, 15\%--50\% of input tokens are randomly masked for prediction. The maximum sequence length is 512 tokens, training uses TF32 precision, and the regime predictor is kept frozen, with its input length capped at 192 timesteps.

For fine-tuning on M5, we use a substantially smaller learning rate of $10^{-5}$ and no learning-rate schedule. The batch size is set between 8 and 16 depending on the configuration, and models are fine-tuned for 6,000--10,000 steps. We apply level-balanced sampling so that each M5 hierarchy level is sampled with equal probability during training. The regime predictor remains active, and the KL alignment loss uses the same coefficient as in pre-training.

\subsection{Feature Correlation Analysis}
To assess whether the proposed structural descriptors provide complementary information, Table~\ref{tab:feature_correlation} reports their Pearson correlation matrix on 98,910 training samples from the pre-training pool. The off-diagonal correlations are uniformly moderate in magnitude, with the largest absolute correlation equal to 0.296. These results suggest that forecastability, seasonality, trend, and sparsity capture distinct temporal properties and therefore provide complementary, non-redundant signals for both the regime predictor and the routing prior.

\begin{table}[ht!]
\centering\small
\caption{Architecture details of \modelname{} variants.}
\label{tab:ame_family}
\begin{tabular}{@{}lccccc@{}}
\toprule
Model & $d$ & $L$ & $E$/$k$ & Params & Active \\
\midrule
tiny   &  256 & 4 & 5/2  &  14M &   6M \\
small  &  256 & 5 & 10/2 &  30M &   7M \\
base   &  512 & 5 & 5/2  &  62M &  29M \\
large  &  768 & 6 & 5/2  & 162M &  75M \\
ultra  & 1024 & 6 & 10/2 & 540M & 133M \\
\bottomrule
\end{tabular}
\end{table}

\begin{table}[ht!]
\caption{Pearson correlation between the four structural descriptors across 98,910 training samples. Moderate correlations ($|r| < 0.5$) indicate that the features capture distinct temporal properties.}
\label{tab:feature_correlation}
\centering
\small
\begin{tabular}{lcccc}
\toprule
 & \textbf{Forecast.} & \textbf{Season.} & \textbf{Trend} & \textbf{Sparsity} \\
\midrule
Forecastability & 1.000 &  0.111 &  0.200 & $-$0.296 \\
Seasonality     & 0.111 &  1.000 & $-$0.280 & $-$0.031 \\
Trend           & 0.200 & $-$0.280 &  1.000 & $-$0.277 \\
Sparsity        & $-$0.296 & $-$0.031 & $-$0.277 &  1.000 \\
\bottomrule
\end{tabular}
\end{table}

\subsection{Additional Results Tables}
Table~\ref{app_tab:main_results} reports the full GIFT-Eval results across all
four metrics, including sMAPE and MAE, and includes the task-specific per-dataset
comparison omitted from the compact main-paper summary in Table~\ref{tab:main_results}.
Additional ablations are reported in Table~\ref{tab:ablation}.
All public datasets and benchmarks are used according to their released licenses and terms.
GIFT-Eval is released under Apache-2.0. We use public datasets from LOTSA, Chronos,
and M5 through their official releases and cite their original sources.
\begin{table}[t]
\caption{Forecasting performance on GIFT-Eval (97 tasks). Scores are the geometric mean of each metric normalized by the Seasonal Naive baseline (lower is better). Best results within each scale group are shown in \textbf{bold}. $^\dagger$ Active parameters per token due to sparse MoE routing.}
\label{app_tab:main_results}
\centering
\small
\begin{tabular}{llccccc}
\toprule
\textbf{Scale} & \textbf{Model} & \textbf{\#Params} & \textbf{MASE $\downarrow$} & \textbf{sMAPE $\downarrow$} & \textbf{RMSE $\downarrow$} & \textbf{MAE $\downarrow$} \\
\midrule
\multirow{3}{*}{\rotatebox{0}{Small}}
 & Moirai-Small \cite{woo2024unified}          & 14M   & 0.946 & 1.109 & 1.056 & 0.906 \\
 & Chronos-Small \cite{ansari2024chronos}         & 46M   &0.892  & 1.059 & 0.866 & 0.847 \\
  & \textbf{\modelname{} Tiny} & \textbf{14M}$^\dagger$\textbf{6M}  & 0.856& 1.043 & 0.808 & 0.834 \\
 & \textbf{\modelname{} Small} & \textbf{30M}$^\dagger$\textbf{7M}  & 0.812& 0.990 & 0.767 & 0.741 \\
\midrule
\multirow{3}{*}{\rotatebox{0}{Base}}
 & Chronos-2    \cite{ansari2025chronos}         & 120M  & 0.698 & 0.886 & 0.704 & 0.679 \\
 & Sundial-128M   \cite{liu2025sundial}       & 128M  & 0.750 & 0.945 & 0.739 & 0.739 \\
 & \textbf{\modelname{} Base} & \textbf{62M}$^\dagger$\textbf{29M} & 0.765 & 0.926 & 0.737 & 0.725 \\
\midrule
\multirow{3}{*}{\rotatebox{0}{Large}}
 & TimesFM-2.5   \cite{das2024decoder}        & 231M  & 0.705 & 0.856 & 0.694 & 0.687 \\
 & Moirai2  \cite{liu2025moirai}             & 300M  & 0.728 & 0.925 & 0.739 & 0.705 \\
 & \textbf{\modelname{} Large}  & \textbf{162M}$^\dagger$\textbf{75M} & 0.700 &  0.852 & 0.690 & 0.680 \\
  & \textbf{\modelname{} Ultra}  & \textbf{540M}$^\dagger$\textbf{133M} & 0.692 & 0.857 & 0.687 & 0.680 \\
 \midrule
\multirow{3}{*}{Per-dataset}
 & PatchTST  \cite{nie2023a}   & --    & 0.849 & 1.020 & 0.839 & 0.810 \\
 & iTransformer \cite{liu2023itransformer} & --    & 0.893 & 1.074 & 0.858 & 0.850 \\
 & \textbf{\modelname{}}  & -- & 0.844 & 0.946 & 0.832 & 0.800 \\
\bottomrule
\end{tabular}
\end{table}

\begin{table}[t!]
\centering
\small
\centering
\caption{Ablation study of \modelname{}{} Tiny on GIFT-Eval benchmark.}
\label{tab:ablation}
\vspace{2mm}
\begin{tabular}{lc}
\toprule
\textbf{Configuration} & \textbf{MASE $\downarrow$} \\
\midrule
\multicolumn{2}{l}{\textit{(a) Architecture}} \\
\textbf{\modelname{} Tiny (proposed)} & \textbf{0.856} \\
Standard MoE (no prior) & 0.929 \\
Dense (no MoE) & 0.958 \\
\midrule
\multicolumn{2}{l}{\textit{(b) Prior integration}} \\
Reverse KL & 0.917 \\
Additive prior & 1.060 \\
\midrule
\multicolumn{2}{l}{\textit{(c) Regime ablation}} \\
Drop Forecastability & 0.919 \\
Drop Seasonality & 0.889 \\
Drop Trend & 0.882 \\
Drop Sparsity & 0.910 \\
\bottomrule
\end{tabular}
\end{table}



\end{document}